%% file: neurips_2026.tex
\documentclass{article}

\PassOptionsToPackage{numbers, compress}{natbib}
\usepackage[preprint]{neurips_2026}

\usepackage[utf8]{inputenc} 
\usepackage[T1]{fontenc}    
\usepackage{hyperref}       
\usepackage{url}            
\usepackage{booktabs}       
\usepackage{amsfonts}       
\usepackage{nicefrac}       
\usepackage{microtype}      
\usepackage{xcolor}         
\usepackage{multirow}
\usepackage{graphicx}
\usepackage{tabularx}
\usepackage{array}
\usepackage{amsmath}
\usepackage{amssymb}
\usepackage{enumitem}
\usepackage{natbib}
\usepackage{subcaption}
\usepackage{wrapfig}
\usepackage[export]{adjustbox}
\usepackage{tcolorbox}
\tcbuselibrary{skins, breakable}

\newcommand{\OurBenchmark}{SLVMBench }
\definecolor{exampleblue}{RGB}{0, 102, 204}

\newtcolorbox{promptbox}[2][]{
  enhanced,
  breakable,
  colback=white,
  colframe=black!70,
  colbacktitle=black!70,
  coltitle=white,
  title={#2},
  fonttitle=\bfseries,
  left=6pt, right=6pt, top=6pt, bottom=6pt,
  boxrule=0.6pt,
  arc=2mm,
  #1
}

\title{SLVMBench: Skill Learning from Video Memory}

\author{%
  Yudong Yang$^1$, Guangzhi Sun$^2$, Yixuan Li$^1$, Chao Zhang$^1$ \\
  $^1$Tsinghua University \quad $^2$University of Cambridge\\
  \texttt{yang-yd21@mails.tsinghua.edu.cn}\quad \texttt{cz277@tsinghua.edu.cn} \\
}

\begin{document}

\maketitle

\begin{abstract}

  We introduce Skill Learning from Video Memory (\OurBenchmark), the first benchmark that jointly evaluates whether video large language models (video-LLMs) can learn skills from long video memory and apply them to real-time tasks. \OurBenchmark presents models with up to 2-hour video streams that contain a tutorial video embedded in a stream of arbitrary irrelevant videos, resembling real-world human learning practices. Video-LLMs are asked to apply the acquired skill to answer real-time questions about an ongoing video. Unlike long-video understanding benchmarks that emphasize passive comprehension and skill-learning benchmarks that rely on short, immediate demonstrations, \OurBenchmark tests the full pipeline of memorizing and extracting procedural knowledge, as well as transferring it to real-time tasks. Moreover, rigorous human annotations feature sub-second-level temporal calibration, manually engineered questions eliminating common-sense guessing, and collated tutorials to ensure coverage of the required skills. Evaluations on state-of-the-art proprietary and open-source video LLMs show that video-LLMs struggle substantially with learning and applying skill knowledge from videos. Moreover, performance degrades markedly when the skill knowledge is placed within a long video memory. These results reveal a key limitation of existing video LLMs and position \OurBenchmark as the first benchmark for studying real-time skill acquisition and application from long-context video memory.
\end{abstract}

\section{Introduction}

Recent advances in audio-visual large language models (video-LLMs) have shown a remarkable ability to understand short videos spanning a couple of minutes, enabling progress in tasks such as captioning, question answering, and multimodal reasoning \cite{li2024llavaov,zhang2024video,wang2024qwen2,lin2024vila,bai2025qwen25vltechnicalreport,damonlpsg2025videollama3,vsalmonn2,vsalmonno1,qwen3vl}. However, when video-LLMs are used to power AI agents in real-world scenarios, accurate real-time perception and effective long-term memory are both indispensable requirements for those models. For instance, when performing a certain operation on software, as humans do, the agent should be able to learn from the demonstration of this task that it has seen a couple of hours ago. The evaluation of such abilities is a critical indicator to expand the scope of application for embodied AI.

\begin{figure}[t]
    \centering
    \includegraphics[width=\textwidth]{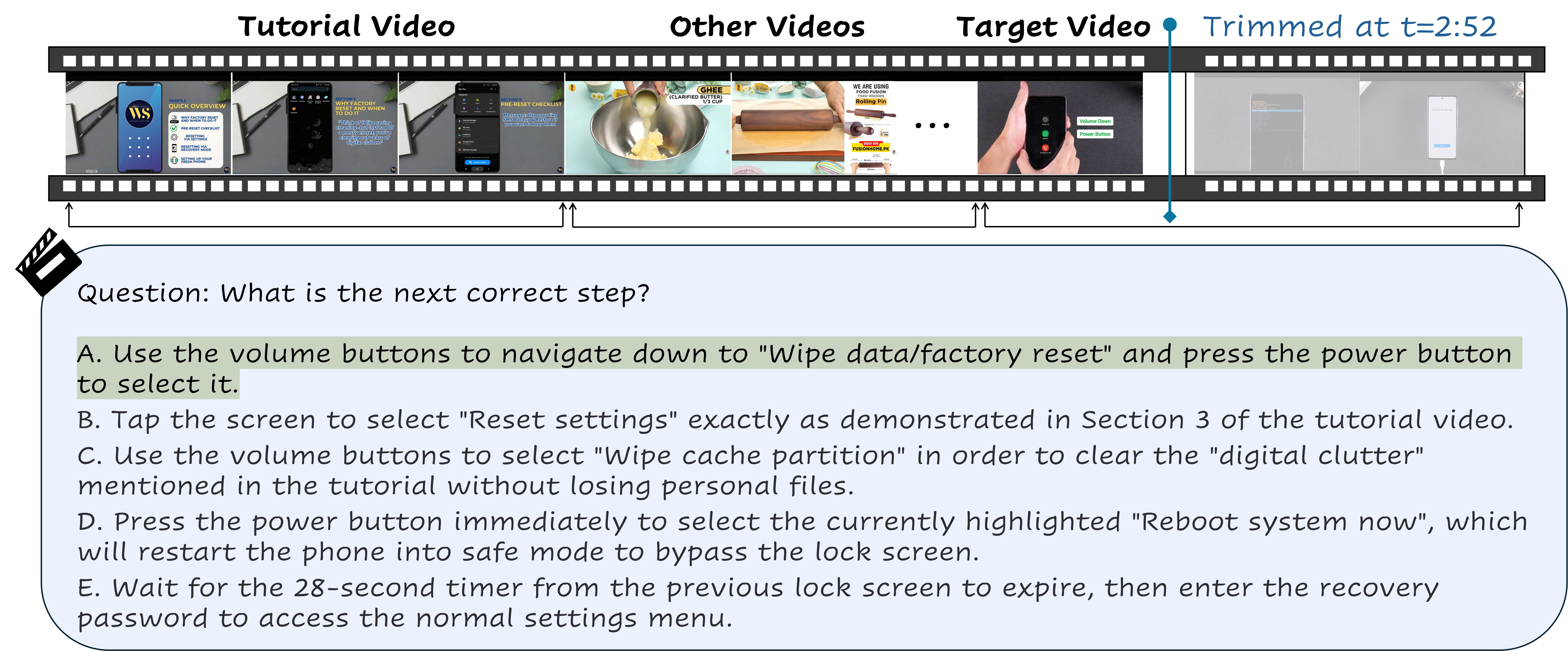}
    \caption{\textbf{Overview of the SLVMBench Task}. The task evaluates multimodal episodic memory by requiring models to: (i) learn procedural knowledge from an initial Tutorial Video; (ii) maintain this knowledge throughout a long sequence of distractor videos; and (iii) perform predictive reasoning in a Target Video at a precise temporal cutoff.} 
    \label{fig:task_figure}  
\end{figure}

A stream of research focuses on improving long-term memory for video-LLMs under streaming settings to perform real-time question answering, using approaches such as KV cache compression \cite{streamingtom,yang2025streammemqueryagnostickvcache,di2025rekv,kim2025infinipot_v,hermes} or the design of memory mechanisms to aggregate tokens \cite{song2024moviechatdensetokensparse,streamforest,flashvstream,vsalmonns}. Despite the progress, one of the key limitations lies in the evaluation of these approaches, which is often separated into two disjoint aspects: To evaluate real-time video perception, benchmarks such as StreamingBench \cite{streamingbench} and OVOBench \cite{ovobench} are widely adopted \cite{yang2025streammemqueryagnostickvcache, hermes, streamforest}, where the video memory required is often limited to 10-20 minutes without requiring long-term memory. On the other hand, to evaluate long video memory capabilities, a series of long-video benchmarks, such as VideoMME \cite{videomme}, LVBench \cite{wang2024lvbench}, and LongVideoBench \cite{longvideobench}, are often used. However, these benchmarks do not require any real-time processing, and instead of assessing whether models can learn procedural knowledge (i.e., skills) from videos, they often primarily measure passive comprehension and information extraction.


To fill the gap, this paper proposes Skill Learning from Video Memory Benchmark (SLVMBench) as the first benchmark that jointly evaluates video-LLMs' ability to learn procedural knowledge from long video memory and apply the knowledge to perform real-time tasks. As shown in Fig.~\ref{fig:task_figure}, each sample in \OurBenchmark comprises one tutorial video and one target video, where the tutorial video is embedded arbitrarily in a stream of 2-3 hours of other irrelevant videos, which serve as the context before the target video. Each question is asked at a specific timestamp, querying about a variety of information such as procedural prediction, tool constraints and diagnostic adaptation, etc., with reference to the real-time ongoing video. The contribution of this paper is summarized as follows.

\begin{itemize}[leftmargin=*]
    \item We propose SLVMBench, the first benchmark jointly evaluating skill acquisition from long video memory and application of acquired skill to real-time tasks. SLVMBench contains 2,261 human-generated questions-answer pairs from 1,220 videos spanning 11 categories.
    \item Instead of directly using existing video datasets, SLVMBench adopts a rigorous human annotation and verification pipeline to collect a new set of videos dated after 2025. Human annotation in SLVMBench provides high-quality question and option generation, sub-second-level timestamp verification, as well as guaranteed skill knowledge coverage of tutorial videos.
    \item SLVMBench is evaluated on a diverse suite of state-of-the-art proprietary and open-source video LLMs. Experimental results reveal substantial limitations in current models’ ability to acquire procedural knowledge and transfer it to real-time tasks. Notably, performance degrades significantly as the temporal distance between the tutorial video and the target task increases.
\end{itemize}


\section{Related Work}
\begin{table}[htbp]
    \centering
    \caption{Comparison of SLVMBench against other benchmarks for long-video and streaming-video understanding. Realtime indicates whether the task requires real-time perception ability. The tutorial video indicates whether a tutorial video teaching the exact skill is present in the video stream. Uploaded time indicates whether videos are filtered based on the time they are uploaded.}
    \vspace{0.2cm}
    \begin{tabular}{lcccc}
    \toprule
    \textbf{Benchmark}     &  \textbf{Typical Duration} & \textbf{Realtime} & \textbf{Tutorial Video} & \textbf{Uploaded Time} \\
    \midrule
    VideoMME (long) \cite{videomme}     & 1 hour & $\times$ & $\times$ & $\times$ \\
    LVBench \cite{wang2024lvbench} & 3 hours & $\times$ & $\times$ & $\times$ \\
    StreamingBench \cite{streamingbench} & 30 mins & \checkmark & $\times$ & $\times$ \\
    DemoICL \cite{demoicl} & 10 mins & \checkmark & $\times$ & $\times$ \\
    VideoMMMU \cite{videommmu} & 8 mins & $\times$ & \checkmark & $\times$ \\ 
    ELViM \cite{vsalmonns} & 1.5 hours & \checkmark & $\times$ & \checkmark \\
    \midrule
    SLVMBench (ours) & 1-2 hours & \checkmark & \checkmark & \checkmark \\
    \bottomrule
    \end{tabular}
    \label{tab:compare}
\end{table}

Our paper is related to long-video understanding benchmarks \cite{wang2024lvbench,longvideobench,lvomnibench}, streaming video understanding benchmarks \cite{streamingbench,ovobench}, and is more closely related to very recent research efforts in skill learning from videos. A more direct and multifaceted comparison with the aforementioned benchmarks is provided in Table~\ref{tab:compare}, motivating the design of SLVMBench.

VideoMMMU \cite{videommmu} is the first work that evaluates the ability of video-LLMs to acquire knowledge from tutorial videos and apply it to questions that are covered by the video but not exact matches. This benchmark focuses predominantly on STEM questions where the videos are lectures and questions are text or text-image-based, without requiring real-time understanding. On the contrary, DemoICL \cite{demoicl} finds paired similar videos from the HowTo100 \cite{howto100} dataset and uses one as the tutorial and another as the target to perform real-time question-answering. However, unlike VideoMMMU, there is no guaranteed knowledge coverage in DemoICL, and hence, the improvement from presenting the tutorial video is somewhat limited. Compared to both benchmarks, SLVMBench focuses on testing skill knowledge from long video memory, hence comprises much longer video context with tutorial videos covering the required knowledge. Moreover, SLVMBench contains a new collection of videos dated after 2025, minimizing the inherent knowledge and emphasizing the contribution from the video memory. In addition, ELViM is the benchmark testing both long-term memory and real-time questions. However, instead of using a tutorial video and evaluating knowledge transfer, it replays the target video in a long video stream, focusing on evaluating the recall of previously seen clips.

\begin{table}[t]
\centering
\small
\caption{Task categories and examples in {\OurBenchmark{}}. All tasks are presented in a 5-option Multiple Choice format, requiring models to combine tutorial knowledge with real-time target video perception.}
\vspace{0.2cm}
\label{tab:task_examples}
\begin{tabularx}{\textwidth}{lc>{\centering\arraybackslash\itshape}X}
\toprule
\textbf{Theme} & \textbf{Task Type} & \multicolumn{1}{c}{\textbf{Example}} \cr
\midrule

\multirow{3}{*}{\shortstack{\textbf{Procedural} \\ \textbf{Mastery}}}
& Next Step Prediction & After this, what is the next step? \\
\cmidrule{2-3}
& Step Ordering & What is the second step of the process? \\
\cmidrule{2-3}
& Subgoal Prediction & What is the purpose of it? \\
\midrule

\multirow{3}{*}{\shortstack{\textbf{Constraints} \\ \textbf{ \& Tool Logic}}} 
& Parameter Recall & In what small increments should you raise it? \\
\cmidrule{2-3}
& Tool Configuration & What tool is used to perform the curved cut? \\
\cmidrule{2-3}
& Safety Check & What should you pay attention to when you do that?  \\
\midrule

\multirow{2}{*}{\shortstack{\textbf{Diagnostics \&} \\ \textbf{Adaptation}}} 
& Mistake Detection & What is the common mistake of that for beginners? \\
\cmidrule{2-3}
& Conditional Branching & What can you do if you want to do it?\\
\bottomrule
\end{tabularx}
\end{table}

\section{SLVMBench}
\subsection{Data Curation Principles}
\textbf{Simulating Multimodal Episodic Memory for Procedural Skill Acquisition.} Humans naturally acquire complex skills by observing audio-visual demonstrations and recalling this procedural knowledge from long-term memory when performing tasks. {\OurBenchmark} is specifically designed to simulate this cognitive process within a multimodal context.  Specifically, {\OurBenchmark} requires models to perceive ``demonstration knowledge'' from an extended audio-visual tutorial stream and store the integrated multimodal cues in episodic memory. This knowledge is then retrieved and applied after a significant temporal gap to support real-time task execution.

\textbf{Evaluating Long-form Streaming Context with Extended Temporal Gaps.} A distinguishing feature of {\OurBenchmark} is its emphasis on long-form video understanding. While existing benchmarks often utilize short clips, {\OurBenchmark} incorporates substantially longer videos, lasting up to \textbf{2 hours}. To rigorously test long-term memory, we ensure a significant interval (at least 30 minutes) between the presentation of relevant knowledge and the subsequent query. By inserting long sequences of unrelated ``distractor'' videos between the tutorial and the target task, {\OurBenchmark} challenges models to maintain a stable memory representation and perform effective retrieval amidst high-density noise.

\textbf{Real-time Perception and Precise Predictive Reasoning.} To reflect the requirements of future real-world agents, {\OurBenchmark} evaluates real-time awareness of task progress. Instead of asking about completed actions, questions in {\OurBenchmark} are posed at precise temporal cutoffs, pausing the video at a critical moment just before a key step is performed. This requires the model to not only understand the historical context and demonstration rules but also to have an ongoing perception of the current state to predict the immediate next action.

\subsection{Dataset Statistics}
\begin{figure*}[t]
  \centering
  \begin{minipage}[c]{0.46\textwidth}
    \centering
    \begin{subfigure}{\linewidth}
      \centering
      \includegraphics[width=\linewidth]{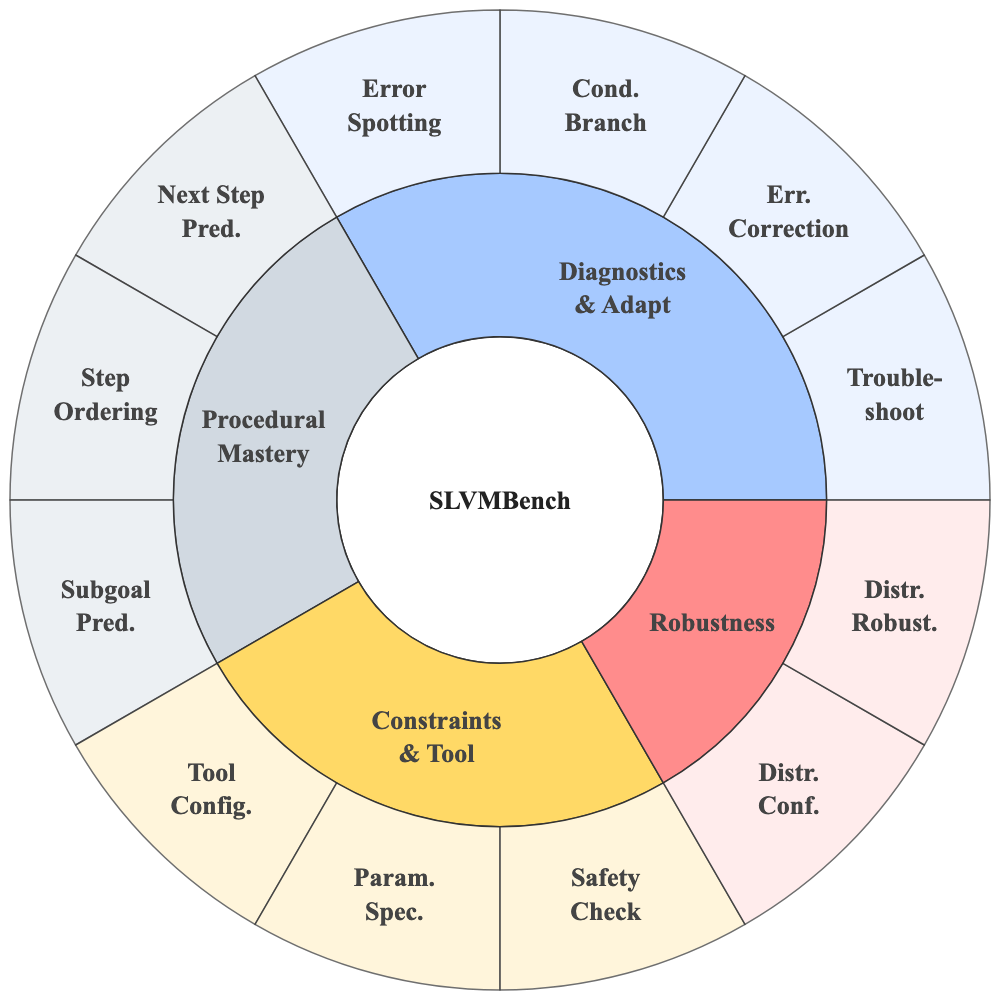} 
      \caption{Hierarchical task categories.}
      \label{fig:task_hierarchy}
    \end{subfigure}
  \end{minipage}
  \hfill
  \begin{minipage}[c]{0.50\textwidth}
    \centering
    \begin{subfigure}{\linewidth}
      \centering
      \includegraphics[width=\linewidth]{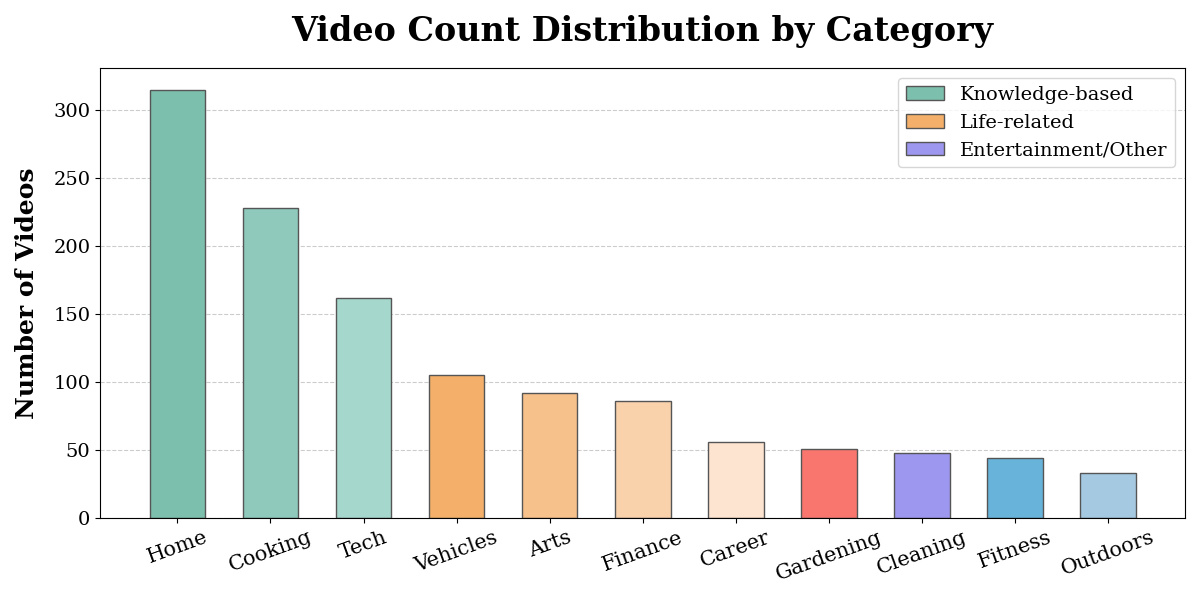}
      \caption{Video count distribution by category.}
      \label{fig:category_dist}
    \end{subfigure}
    
    \vspace{0.8em} 
    
    \begin{subfigure}{\linewidth}
      \centering
      \includegraphics[width=\linewidth]{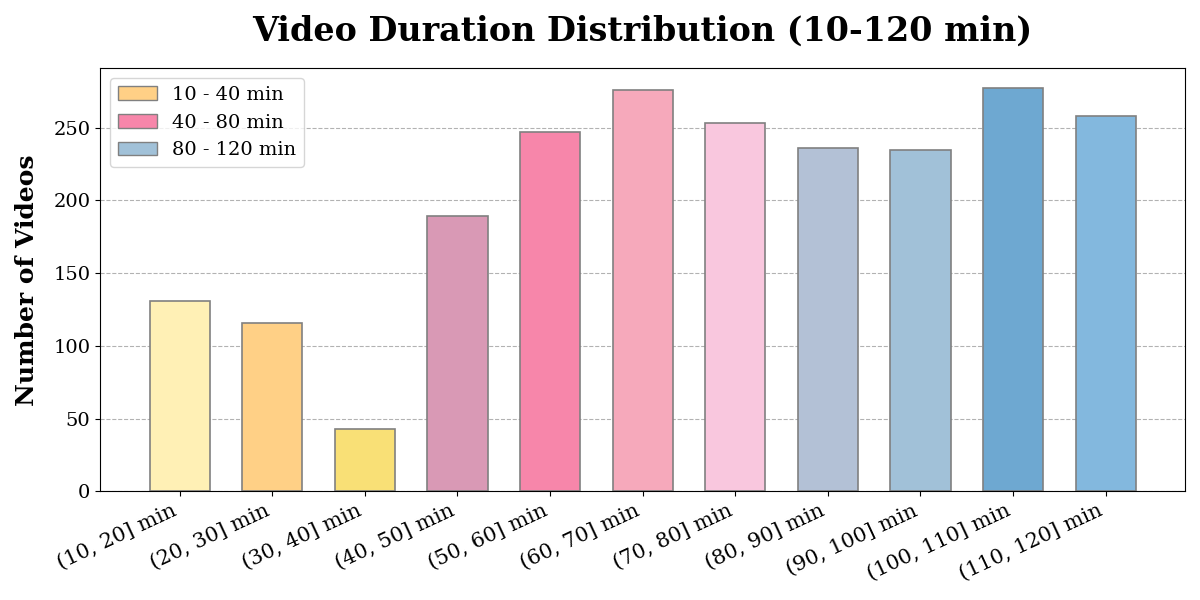}
      \caption{Statistical distribution of video durations.}
      \label{fig:duration_dist}
    \end{subfigure}
  \end{minipage}

  \caption{\textbf{In-depth temporal and categorical analysis of SLVMBench.}
  (a) The hierarchical taxonomy of tasks including procedural mastery and diagnostics. 
  (b) Distribution of videos across 11 major categories. 
  (c) Coverage of video durations from 10 to 120 minutes, ensuring long-horizon evaluation.}
  \label{fig:combined_overall}
\end{figure*}

As illustrated in Fig.~\ref{fig:combined_overall}, {\OurBenchmark} provides a diverse and balanced distribution across tasks, domains, and temporal scales. Specifically, our hierarchical taxonomy (Fig.~\ref{fig:task_hierarchy}) encompasses 11 fine-grained task types for multi-dimensional reasoning, while the dataset spans 11 real-world domains ranging from daily life to professional skills (Fig.~\ref{fig:category_dist}). Crucially, the extensive temporal coverage from 10 to 120 minutes (Fig.~\ref{fig:duration_dist}) provides a rigorous testbed for evaluating long-horizon episodic memory and knowledge retention.

\subsection{Task Formulation and Taxonomy}
\label{task definition}
To provide a rigorous framework for evaluating episodic procedural reasoning, we formalize the core task of {\OurBenchmark} as follows. Unlike traditional video QA where the context is static and localized, {\OurBenchmark} requires a model to perform cross-video knowledge transfer and adaptation under long-term temporal interference.

\textbf{Formal Task Definition.} We define an evaluation instance in {\OurBenchmark} as a quintuple $\mathcal{T} = (\mathcal{V}_\text{tut}, \mathcal{D}, \mathcal{V}_\text{tar}, t_c, \mathcal{Q})$, where $\mathcal{V}_\text{tut}$ represents the tutorial video providing the ground-truth procedural demonstrations, $\mathcal{D} = \{d_1, d_2, ..., d_n\}$ is a sequence of distractor videos unrelated to the target task, $\mathcal{V}_\text{tar}$ denotes the target video representing a real-world task execution scenario, $t_c$ signifies a precise temporal cutoff (the decision point) within $\mathcal{V}_\text{tar}$, and $\mathcal{Q}$ is a multiple-choice question with five options $\mathcal{O} = \{o_1, ..., o_5\}$. 

The model $f$ is provided with a continuous multimodal stream $\mathcal{S}$ formed by the concatenation of these elements:
\begin{equation}
    \mathcal{S} = \mathcal{V}_\text{tut} \oplus \mathcal{D} \oplus \mathcal{V}_\text{tar}^{[0, t_c]}
\end{equation}
where $\mathcal{V}_\text{tar}^{[0, t_c]}$ denotes the segment of the target video from its start to the cutoff point $t_c$. The objective of the model is to predict the correct answer $a^* \in \mathcal{O}$ by retrieving relevant procedural cues from its episodic memory of $\mathcal{V}_\text{tut}$ and applying them to the current state observed at $t_c$:
\begin{equation}
    \hat{a} = \text{argmax}_{a \in \mathcal{O}} P(a \mid \mathcal{V}_\text{tut}, \mathcal{D}, \mathcal{V}_\text{tar}^{[0, t_c]}, \mathcal{Q})
\end{equation}

Crucially, this setup evaluates the model's episodic memory capability within an extensive long-form video context, requiring the retrieval and application of knowledge across a temporal gap, ranging from 10 minutes to 2 hours. 

\textbf{Fine-grained Task Taxonomy.} To comprehensively evaluate the efficacy of procedural understanding from multiple perspectives, we categorize the designed questions in {\OurBenchmark} into three primary thematic clusters, as illustrated in Table~\ref{tab:task_examples}:

\begin{itemize}[leftmargin=*]
    \item \textbf{Procedural Mastery:} This cluster evaluates sequential logic through \textit{Next Step Prediction} (identifying the immediate subsequent action), \textit{Step Ordering} (sequencing multiple upcoming steps), and \textit{Subgoal Prediction} (identifying the underlying objective behind a specific action).
    
    \item \textbf{Constraints \& Tool Logic:} This dimension assesses the model's ability to recall and apply specific technical requirements and tool-related principles. It includes \textit{Parameter Recall} (retrieving specific values like increments or settings), \textit{Tool Configuration} (identifying the correct tool or mode for a task), and \textit{Safety Checks} (determining critical conditions or precautions).
    
    \item \textbf{Diagnostics \& Adaptation:} Requiring higher-order cognitive flexibility, this theme evaluates how models handle deviations and procedural complexity. It encompasses \textit{Mistake Detection} (identifying common errors or deviations from the tutorial) and \textit{Conditional Branching} (adapting to specific scenarios or ``if-then'' logic within the procedure).
\end{itemize}

\subsection{Data Construction Pipeline}
\begin{figure}[t]
    \centering
    \includegraphics[width=\textwidth]{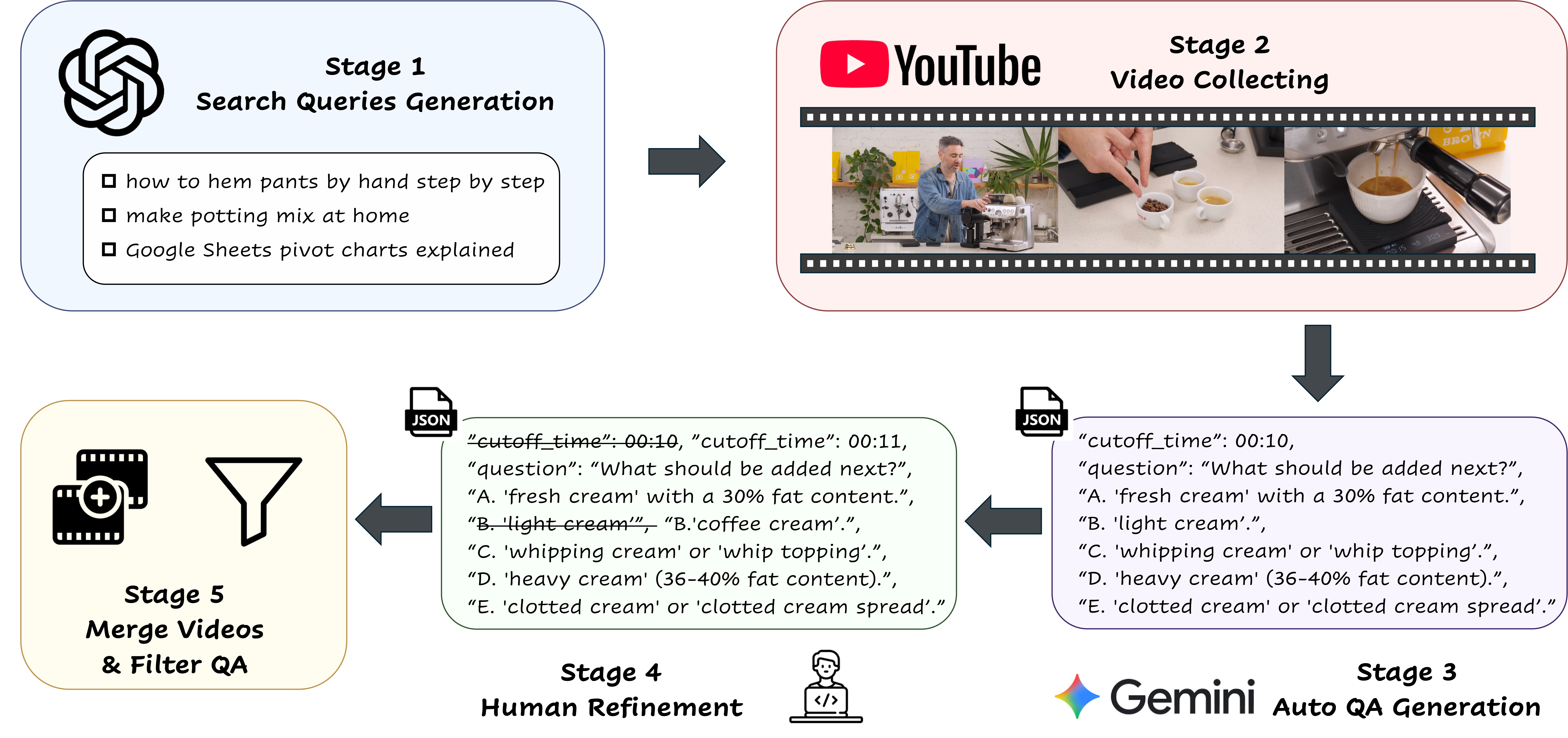}
    \caption{\textbf{Overview of the multi-stage SLVMBench data construction pipeline.} (Stage 1-2) GPT-4o generates a wide array of diverse procedural queries to crawl high-quality instructional content from YouTube. (Stage 3) Gemini-2.5-Pro is employed for automated QA synthesis and initial temporal anchoring based on the video content. (Stage 4) Human experts meticulously refine the generated JSON data to calibrate timestamps with sub-second precision and modify ``spoiler'' options to ensure logical rigor. (Stage 5) Final videos are merged with long-duration distractor sequences and filtered to establish the high-fidelity streaming benchmark.}
    \label{fig:pipeline}  
\end{figure}

The construction of {\OurBenchmark} follows a structured pipeline. The process is divided into five stages, as illustrated in Fig.~\ref{fig:pipeline}.

\textbf{Procedural Query Generation and Video Acquisition.} We employ GPT-4o to generate diverse procedural and ``how-to'' search queries across multiple domains to crawl instructional content from YouTube. Using these queries, we crawl a large corpus of candidate videos from YouTube. Crucially, we restrict our collection to videos published after January 2025 to mitigate data contamination and ensure models rely on reasoning rather than memorized training data.

\textbf{Manual Pairing of Tutorial and Target Videos.} The core of our pipeline is the manual selection and verification of Tutorial-Target video pairs. Annotators confirm that each tutorial provides the specific rules or instructions necessary to solve the target task, ensuring a direct logical link even when visual environments differ. This human-verified connection ensures {\OurBenchmark{}} evaluates true knowledge transfer instead of simple visual pattern matching.

\textbf{Automated QA Synthesis and Temporal Anchoring.} After establishing the video pairs, we utilize Gemini-2.5-Pro to generate automated QA candidates. Guided by our task taxonomy (defined in Section~\ref{task definition}), Gemini analyzes the video streams to identify critical \textbf{decision points}: moments in the target video where a subsequent action is contingent on the tutorial's instructions. The model then generates a question, a correct answer, and five plausible distractors, while simultaneously predicting a precise temporal cutoff.

\textbf{Manual Fine-grained Annotation.} To ensure high-fidelity data, all 2,261 samples in {\OurBenchmark{}} underwent rigorous manual annotation and auditing, requiring an average of 20 minutes per instance (over \textbf{750 total man-hours}). The refinement focuses on three dimensions: (1) Consistency and Validity: Annotators verify the logical alignment between tutorial and target videos, rejecting pairs where instructional cues are insufficient for task execution. (2) Precise Temporal Anchoring: Video cutoffs are manually calibrated to sub-second precision, ensuring the stream is paused at a precise decision point immediately after a previous step ends but before the next action begins. (3) Anti-Guessing Design: Questions are ``de-spoiled'' to hide inferable info, and distractors are manually engineered to be ``plausible but incorrect''. These measures eliminate reliance on common-sense priors, forcing the model to rely exclusively on its episodic memory. Detailed information regarding annotator demographics, training protocols, and specific labeling logistics is provided in the Appendix~\ref{human_annotation}.

\section{Experiment}

\subsection{Evaluation Settings}
\input{result_all}
\textbf{Models.} We evaluate a representative suite of both proprietary and open-source MLLMs to establish robust baselines on {\OurBenchmark}. For proprietary models, we select Gemini 3.1 Pro \cite{gemini3_2025}, GPT-5.2 \cite{singh2025openaigpt5}, and GPT-4o \cite{hurst2024gpt4o} to assess the performance ceiling in long-context episodic reasoning. For open-source models, we evaluate specialized streaming frameworks and memory-augmented architectures, including video-SALMONN-S \cite{vsalmonns}, StreamMem \cite{streammem}, and PEMF \cite{pemf}, alongside non-streaming architectures such as video-SALMONN 2+ (8B) \cite{vsalmonn2}, Qwen3-Omni (30B) \cite{Qwen3-Omni}, Qwen2.5-Omni (7B) \cite{Qwen2.5-Omni}, VideoLLaMA3 (7B) \cite{damonlpsg2025videollama3}, and Qwen3-VL (8B) \cite{qwen3vl}. Note that video-SALMONN 2+ uses Qwen3-VL (8B) backbone and finetuned on the dataset provided in \cite{vsalmonn2}. StreamMem and PEMF use video-SALMONN 2+ as the backbone and directly perform compression. For open-source models, all inference experiments were conducted on a single H800 GPU.

\textbf{Metrics.} Regarding the evaluation metric, we adopt \textbf{Accuracy} as the primary measure. Since each task is structured as a five-option multiple-choice question, Accuracy is calculated by directly matching the model's predicted choice with the ground-truth option. Detailed implementation details and specific test configurations for each model are provided in the Appendix~\ref{model configurations}. In addition to accuracy, we also report Wilson confidence intervals with details described in Appendix~\ref{sec:ci}.

\textbf{Evaluation Paradigms and Configurations.} To systematically investigate the roles of knowledge acquisition and episodic memory, we design three distinct evaluation paradigms: (1) \textbf{w/o Tutorial}, where only the target video truncated at the decision point is provided. This setting serves as a baseline to measure the model's reliance on common-sense priors without any instructional guidance. (2) \textbf{w/ Tutorial}, where the model receives a merged video consisting of the complete tutorial followed immediately by the truncated target. This assesses the model's ability to promptly apply learned procedural knowledge. (3) \textbf{w/ Tutorial (Long)}, where a long sequence of unrelated distractor videos is inserted between the tutorial and the target to create an extended temporal gap. The resulting merged videos are uniformly distributed in length, ranging from 10 minutes to 2 hours with a 10-minute increment per level. This paradigm evaluates the robustness of the model's episodic memory and its capability to retrieve relevant cues amidst dense information noise over time. For all settings, we strictly follow the official configurations of each model for frame sampling and hyperparameter settings to ensure fairness. The prompts are kept minimal, containing only the specific question and options without additional instructional templates, thereby focusing exclusively on the model's inherent reasoning capabilities.

\subsection{Quantitative Results}
\input{task_result}

The main evaluation results on the {\OurBenchmark} are presented in Table~\ref{tab:main_results}. By analyzing the performance across different paradigms, we derive the following observations regarding the validity of the benchmark and the capabilities of current MLLMs.

\textbf{Benchmarking Validity and Task Difficulty.} The results first validate the fundamental design of {\OurBenchmark}. Comparing the \textit{w/o Tutorial} (baseline) to the \textit{w/ Tutorial} setting, we observe a significant and consistent performance gain ($\Delta$) across all models. For instance, Gemini 3.1 Pro and video-SALMONN 2+ (768 frames) exhibit substantial improvements of 29.47\% and 29.72\%, respectively. This large gap proves the rationality of our dataset: the tasks are designed such that the tutorial provides essential procedural knowledge that is otherwise unavailable through common-sense priors. Simultaneously, the benchmark presents an extreme difficulty in episodic reasoning. When a 2-hour temporal gap is introduced in the \textit{w/ Tutorial (Long)} setting, nearly all models suffer from a severe performance ``cliff''. Most notably, the performance gain over the baseline ($\Delta_{L}$) for proprietary models like GPT-5.2 and GPT-4o shrinks to only 2.21\% and 2.48\%, respectively. Similarly, open-source models such as Qwen3-Omni and VideoLLaMA3 show minimal improvements of 2.33\% and 0.64\%, almost reverting to their zero-tutorial baseline. This widespread performance decay underscores that maintaining high-fidelity episodic memory amidst long-form distractors remains a major unsolved challenge for state-of-the-art MLLMs.

\textbf{Impact of Multimodal Cues: Audio-Visual vs. Visual-Only Models.} We observe that audio integration is a decisive factor for acquiring procedural knowledge, as critical instructions in tutorials are often delivered verbally. In the proprietary tier, the audio-visual model Gemini 3.1 Pro significantly outperforms the visual-only model GPT-5.2, with its episodic retention ($\Delta_L=14.00$) far exceeding GPT-5.2's ($\Delta_L=2.21$). Crucially, even at a smaller 7B/8B parameter scale, the audio-visual model Qwen2.5-Omni (7B) exhibits remarkable memory robustness ($\Delta_L=11.2$), whereas the visual-only model Qwen3-VL (8B) almost completely loses its tutorial-derived gains over a long horizon ($\Delta_L=1.28$). Furthermore, we observe that the density of visual information also contributes to performance; our comparative study with video-SALMONN 2+ shows that increasing the frame count from 64 to 768 significantly bolsters episodic retention, raising $\Delta_L$ from 9.60 to 15.79. These results demonstrate that while processing verbal cues is essential for resolving ambiguities in demonstrations, high-resolution temporal sampling further stabilizes the acquisition and retrieval of high-fidelity procedural memory.

\textbf{Performance of Specialized Streaming and Memory Architectures.} {\OurBenchmark{}} reveals a significant advantage for models with specialized streaming or memory-augmented architectures when handling extended temporal gaps. While standard non-streaming models—including proprietary giants like GPT-4o and GPT-5.2 exhibit ``memory collapse'' with $\Delta_L$ values dropping to near-baseline levels, streaming frameworks such as video-SALMONN-S demonstrate better robustness, maintaining $\Delta_L$ scores of 14.95. However, the inconsistent performance across the streaming category is noteworthy; for instance, PEMF experiences a dramatic accuracy plunge from 64.88\% to 39.36\% ($\Delta_L = 2.39$) in the long-form setting. This disparity suggests that simply extending the input window or processing at a constant 1 FPS is insufficient for effective episodic reasoning with merging or KV-cache selection-based streaming methods. Instead, these results underscore that true episodic retrieval requires sophisticated architectural mechanisms capable of selectively filtering information noise to access relevant procedural cues after a 2-hour distractor sequence.

\subsection{Fine-grained Task Analysis}
\textbf{Table~\ref{tab:Task result} provides a detailed performance breakdown across eight sub-task categories.} Notably, the state-of-the-art model Gemini 3.1 Pro continues to demonstrate a balanced performance profile, with accuracies ranging from 56.19\% (CB) to 61.18\% (TC). This consistency across cognitive dimensions suggests that {\OurBenchmark{}} establishes a robust and high-quality challenge, where no single sub-task serves as a trivial shortcut.

\textbf{Experimental results reveal a significant advantage for specialized \textbf{streaming architectures} in logic-intensive tasks.} In categories requiring high-level reasoning such as \textit{Mistake Detection} (MD) and \textit{Conditional Branching} (CB), models like video-SALMONN-S achieves remarkable results, with accuracies reaching 73.16\% in CB. This significantly outperforms proprietary models like GPT-4o, which plunges to 39.53\% in CB and a mere 30.61\% in MD. These results suggest that the continuous temporal tracking inherent in streaming models is crucial for identifying execution errors and adaptive branching logic over a long horizon.

\textbf{In contrast, non-streaming models exhibit a clear bottleneck in diagnostic reasoning.} Non-streaming models like Qwen3-VL struggle with diagnostic reasoning, dropping to 16.09\% in CB, which reveals a lack of episodic memory necessary for deep procedural intelligence. GPT-4o’s failure in Mistake Detection (30.61\%) further underscores that identifying procedural deviations remains a significant unsolved challenge for general-purpose MLLMs.

\subsection{Memory Decay Analysis}

\begin{wrapfigure}{r}{0.6\textwidth} 
  \centering
  \includegraphics[width=0.58\textwidth]{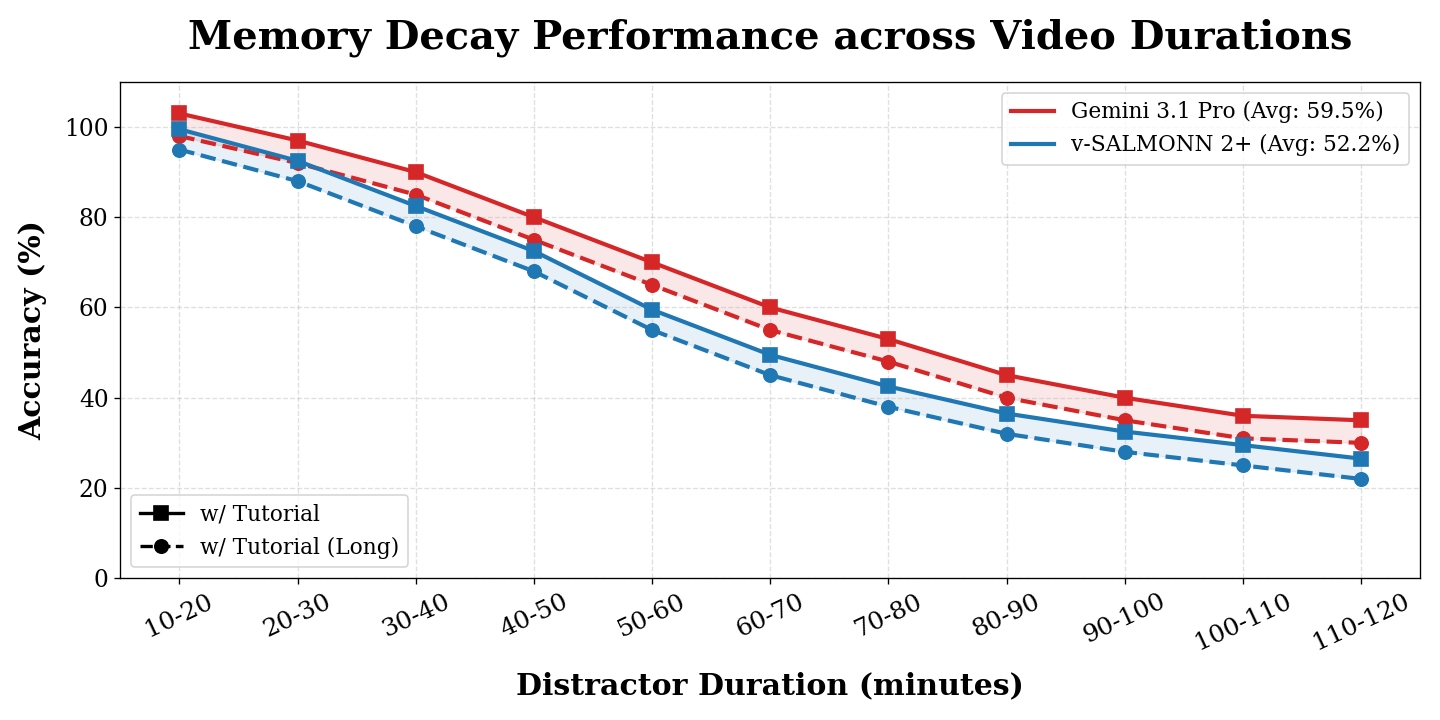}
  \caption{\textbf{Memory decay analysis.} Accuracy decay curves for Gemini 3.1 Pro and video-SALMONN 2+.}
  \label{fig:memory_decay}
\end{wrapfigure}

As illustrated in Fig.~\ref{fig:memory_decay}, the accuracy of the model is inversely correlated with the duration of the distractor in all architectures evaluated. The decay curves reveal distinct cognitive thresholds: while performance remains relatively stable during the first hour for the proprietary model, a sharp ``memory cliff'' is observed for the open-source counterpart after just 30 minutes of interference. This large degradation underscores a major bottleneck in long-term video memory, justifying the necessity of our 2-hour evaluation window to identify the true limits of streaming AI agents.

Comparing the two architectures, Gemini 3.1 Pro (red) maintains a higher overall performance ceiling, yet follows a similarly steep downward trajectory as video-SALMONN 2+ (blue). The shaded region between the lines represents the ``Memory Gap'', the performance loss specifically attributed to the episodic distance between the tutorial and the target task. The fact that precision drops significantly as the temporal gap approaches two hours justifies the need for the long-horizon evaluation design of {\OurBenchmark{}} to identify the true reasoning limits of streaming AI agents.

\section{Limitations}

This work bears the following two limitations: (i). The temporal difference between the tutorial and the application timestamp is limited to up to 2 hours. We compromise on this design due to the limitation of input sequence lengths in current video-LLMs. (ii). While this is intended to simulate real-world applications for embodied AI, we abstract such interaction into a question-answering format rather than in a real-world or simulated environment for AI agents. Future work will explore video demonstration as memory for agentic AI systems. 

\section{Conclusion}

We propose SLVMBench, the first benchmark that jointly evaluates the ability of video-LLMs to learn skills from long video memory, and applies it to real-time question-answering tasks. In SLVMBench, each sample requires the model to watch a long video stream of 2-3 hours up to a certain timestamp where a question is asked, and a tutorial is provided in this long video stream, 10 minutes to 2 hours away from the current timestamp. Rigorous human annotations were provided to generate and verify the tutorial video, the target video and the question, ensuring full coverage of the skill knowledge as well as calibrating the timestamp at sub-second precision. Experiments across a range of state-of-the-art streaming and non-streaming models and approaches have been investigated, demonstrating the challenge that lies in both skill knowledge acquisition and application, and the long-term memory decay in video-LLMs.

{\small
\bibliographystyle{unsrt}
\bibliography{neurips_2026}
}


\appendix
\section{Instruction for Annotators}

All of the following blocks and instruction screenshots are provided to the annotators.

\begin{promptbox}[label={real_scripts}]{Data Background and Quality Control Objectives}
1. \textbf{Data Background}\\
We currently use LLMs to automatically generate a set of streaming QA data based on “tutorial video + target video” pairs. Each data instance contains a Tutorial Video, a Target Video, and predictive/diagnostic QA generated for specific timestamps in the Target Video.

2. \textbf{Quality Control Objectives}\\
Revise each LLM-generated data instance to:

Remove hallucinated data generated by the LLM, such as cases where the text does not match the video or the timestamp is misaligned.
Ensure that each question satisfies the following criterion: without watching the rules in the Tutorial Video or observing the current state in the Target Video, it is absolutely impossible to answer correctly based on common sense alone.
Refine the question stems and answer options so that they better align with natural human language and realistic testing scenarios.
\end{promptbox}

\begin{promptbox}[label={real_scripts}]{Annotation Template Explanation}

\begin{verbatim}
{
    // Tutorial Video, cover the knowledge required by the question
    "orig_video_path": ".../Ft9qhSnF5gE.mp4",
    // Target Video: real execution scenario
    "video_path": ".../qpHWEqJVOl8.mp4",     
    // Topic/search query used to generate this data
    "search_query": "fix leaky faucet cartridge after:2025-01-01", 
    // Total duration of the target video
    "duration_sec": 38.0,                     
    "question": {
        // Key action start time
        "start": 11,
        // Cutoff timestamp
        "end": 13,
        // Original discarded question
        "question": "After removing the faucet handle, 
        what is the next step to access the cartridge?",
        "answer": "Loosen the packing nut with a wrench and then 
        remove the cartridge."
    },
    // Final question after LLM refinement, with spoilers removed
    "modified_question": "Now, what is the next step to 
    access the cartridge?", 
    // Reason for modifying the question, for reference
    "modify_reason": "...",
    "options": [
        // Five candidate options
        "Loosen the packing nut with a wrench and then remove the 
        cartridge.",
        "Remove the bonnet nut covering the cartridge.",
        "Loosen and remove the screw under the cap to detach the handle.",
        "Unscrew the retaining screw hidden under the handle cap to lift
        off the handle.",
        "Unscrew the screw located under the handle cap to take off 
        the handle."
    ],
    // Correct option
    "correct_choice": "A"
}
\end{verbatim}
\end{promptbox}

\begin{promptbox}[label={real_scripts}]{Quality Control Procedure}
\textbf{Step 1: Dual-Video Consistency Check}\\
Operation: Quickly review \texttt{orig\_video\_path} (Tutorial Video) and \texttt{video\_path} (Target Video).
\begin{itemize}
    \item Check: Is the tutorial valid? Does the Tutorial Video clearly demonstrate or verbally explain the standard procedure, steps, or principles for the operation being performed in the Target Video?
    \item Rejection criterion: If the two videos are unrelated, or if the Tutorial Video does not mention the key step in the Target Video, discard the data instance directly.
\end{itemize}

\textbf{Step 2: Precise Timestamp Check}\\
Operation: Open the Target Video (\texttt{video\_path}) and pause strictly at question.end (e.g., the 13-second mark).
\begin{itemize}
    \item Check 1: Is the cutoff point reasonable? When paused at 13 seconds, is the video exactly at a key decision point?
    \item Check 2: Does it reveal the answer? Watch 1–2 seconds after end to ensure that the correct action has not yet occurred.
    \item Modification: If the timestamp is slightly off by 1–3 seconds, manually adjust start and end so that end becomes an ideal suspenseful cutoff point.
\end{itemize}

\textbf{Step 3: Question and Option Logic Check} \\
Operation: Assume you are a test taker who has not seen the subsequent frames and can only rely on the Tutorial Video knowledge plus the first 13 seconds of the Target Video.

\begin{itemize}
    \item Question check (\texttt{modified\_question})
Ensure that the LLM’s refinement is reasonable. Replace potentially spoiler-revealing terms with referring expressions where needed, and make sure the revised question reads naturally and clearly refers to the video context.
\item Correct option check (\texttt{correct\_choice})
Verify that the correct option strictly follows the Tutorial Video and naturally continues from the Target Video state at 13 seconds. If it conflicts with the tutorial, revise the answer or reject the data.
\item Distractor option check (\texttt{options})
Remove obviously implausible distractors, avoid overlapping or ambiguous options, and ensure the options cannot be solved by blind guessing. The best distractors should appear plausible without the tutorial but be incorrect under the task-specific rules. Revise low-quality distractors so the five options describe five distinct, reasonable steps, with only one uniquely correct answer.
\end{itemize}
    
\end{promptbox}

\begin{promptbox}[label={real_scripts}]{Delivery and Annotation Result Format}
For each LLM-generated data instance, annotators should submit the following quality-control results:

1. \textbf{Quality-Control Decision} (\texttt{Status})
\begin{itemize}
    \item Pass: Perfect; no modification is needed.
    \item Modified: The data is usable, but the timestamp, question stem, or options have been manually adjusted.
    \item Reject: The data is fundamentally invalid, such as mismatched image-text content or forced associations, and should be discarded.
\end{itemize}

2. \textbf{Modification Record} (\texttt{Modified\_Fields})
Fill in this field only when Status is Modified. 
\begin{itemize}
    \item Record which fields were modified, e.g., \texttt{end\_time}: 13 -> 11, options: replaced options C and D.
\end{itemize}

3. \textbf{Rejection Reason} (\texttt{Reject\_Reason})
Fill in this field only when \texttt{Status} is Reject.
\begin{itemize}
    \item Briefly describe the reason for rejection, e.g., ``The tutorial video is about changing a tire, while the target video is about repairing a chain; they are completely mismatched." / ``Option A can be selected based on common sense alone without watching the tutorial video."
\end{itemize}

\end{promptbox}

\begin{promptbox}[label={real_scripts}]{Example Cases}

\textbf{Case 1: Pass}
\begin{verbatim}
{
    {
        "orig_video_path": "yBnqC-Xluyw.mp4",
        "video_path": "B1LYXX9Pc58.mp4",
        "search_query": "dishwasher not draining troubleshoot 
        after:2025-01-01",
        "video_id": "yBnqC-Xluyw",
        "duration_sec": 31.0,
        "question": {
            "question": "If the drain pump is not the cause of the 
            dishwasher's draining issue, what is the other possible 
            problem component mentioned in the video?",
            "answer": "The drain hose could also be the 
            cause of the issue.",
            "start": 13,
            "end": 17
        },
        "result": "yes",
        "options": [
            "The motor may not be functioning properly.",
            "A clogged filter could be causing the draining issue.",
            "The dishwasher's control board may be defective.",
            "The drain hose could also be the cause of the issue.",
            "The water inlet valve could be faulty."
        ],
        "correct_choice": "D",
        "modified_question": "If the drain pump is not the cause of the 
        dishwasher's draining issue, what is the other possible 
        problem component mentioned in the video?",
        "modify_reason": "The question is specific and requires the 
        viewer to recall information presented in the video. 
        The details provided, such as \"drain pump\" and 
        \"dishwasher's draining issue,\" are necessary to 
        frame the question clearly and are not redundant 
        descriptions of the video's content at the given 
        timestamp. Therefore, no modification is needed."
    },
}
\end{verbatim}

\textbf{Case 2: Modified - Wrong timestamp}
\begin{verbatim}
{
    {
        "orig_video_path": "8y-5-sqFsuc.mp4",
        "video_path": "pCdZgO-o8Bw.mp4",
        "search_query": "use a headlamp properly outdoors 
        after:2025-01-01",
        "video_id": "8y-5-sqFsuc",
        "duration_sec": 101.0,
        "question": {
            "question": "If you want to access the dimmest possible 
            white light directly from the off state, what should 
            you do?",
            "answer": "Double press the Power Button to directly 
            activate the ULTRALOW mode.",
            "type": "Next Step",
            "start": 39,
            "end": 44
        },
        "result": "yes",
        "options": [
            "Press and hold the Power Button for three seconds
            to directly activate the dimmest light.",
            "Turn the headlamp on by long-pressing the Power Button, 
            then press and hold the Power Button again until 
            the dimmest light is activated.",
            "Short-press the Power Button multiple times immediately 
            after turning it on to cycle through brightness levels 
            and reach the LOW setting.",
            "Double press the Power Button to directly activate the 
            ULTRALOW mode.",
            "Long-press the Power Button for one second to turn the 
            light on, then short-press the Power Button repeatedly 
            until you reach the LOW setting."
        ],
        "correct_choice": "D",
        "modified_question": "If you want to access the dimmest possible 
        white light directly from the off state, what should you do?",
        "modify_reason": "The question is specific and does not contain 
        information that can be easily inferred from the video at the 
        given timestamp. The video is demonstrating the TURBO mode, 
        while the question asks about the ULTRALOW mode from an off state. 
        The details are necessary to specify the action in question."
    },
}
\end{verbatim}

More cases are provided. We omit here for simplicity.
\end{promptbox}

\section{Model Evaluation Details}
\subsection{Implementation and Computing Infrastructure}
All experiments for open-source models were conducted on a high-performance computing cluster equipped with 8 $\times$ NVIDIA H800 (80GB) GPUs. We utilized the \texttt{vLLM} \cite{kwon2023efficient} or \texttt{HuggingFace Transformers} \cite{wolf-etal-2020-transformers} library for inference, depending on the model's official support. For proprietary models, we accessed the latest available API endpoints as of March 2025.

\subsection{Model-Specific Configurations}
\label{model configurations}
As shown in Table~\ref{tab:main_results} of the main paper, we evaluated all models using their official inference configurations or recommended API versions. The specific versions and sampling settings are as follows:

For all evaluations, we set the \textbf{temperature = 0} to ensure deterministic and reproducible outputs. Following standard practices, we employed the \textbf{greedy decoding} strategy as the default sampling protocol for all models to minimize experimental variance.

\textbf{Proprietary Models:}
For proprietary models, we accessed the specific API snapshots to ensure consistency during our evaluation period:

\begin{itemize}[leftmargin=1.5em]
    \item \textbf{Gemini 3.1 Pro}: We used the version identifier \texttt{gemini-3.1-pro-preview}. The model was evaluated at a constant sampling rate of \textbf{1 fps}, allowing it to ingest the full long-horizon video stream natively.
    \item \textbf{GPT-5.2}: We used the specific version \texttt{gpt-5.2-2025-12-11}. This model was evaluated by sampling \textbf{64 uniform frames} across the concatenated video stream.
    \item \textbf{GPT-4o}: We used the version \texttt{gpt-4o-2024-05-13}. Similar to GPT-5.2, it was evaluated using a \textbf{64-frame uniform sampling} strategy.
\end{itemize}

\textbf{Open-Source Streaming Models (Memory-Augmented):}
\begin{itemize}[leftmargin=1.5em]
    \item \textbf{video-SALMONN-S, StreamMem, and PEMF}: These models utilize specialized memory mechanisms and were evaluated at \textbf{1 fps}. This allows the architectures to process the distractor sequence sequentially, mimicking a real-world streaming scenario without exceeding memory limits.
\end{itemize}

\textbf{Open-Source Non-Streaming Models:}
\begin{itemize}[leftmargin=1.5em]
    \item \textbf{video-SALMONN 2+}: Evaluated with \textbf{768 uniform frames and 64 uniform frames}, leveraging its high-frame processing capacity to capture visual details across the 2-hour window.
    \item \textbf{Qwen3-Omni, Qwen2.5-Omni, VideoLLaMA3, and Qwen3-VL}: These models were evaluated using \textbf{64 uniform frames}, aligned with their standard training and evaluation protocols for non-streaming video tasks.
\end{itemize}

\subsection{Evaluation System Prompt}

To ensure consistent and standardized evaluation across all models, we utilized the following zero-shot system prompt. This prompt is designed to strictly enforce a single-letter output format while emphasizing the predictive nature of the task at the video's cutoff point.

\begin{tcolorbox}[
    colback=gray!5, 
    colframe=gray!50, 
    boxrule=0.5pt, 
    arc=2pt, 
    left=10pt, 
    right=10pt, 
    top=10pt, 
    bottom=10pt,
    title=\textbf{System Prompt for SLVM Evaluation}
]
\small
\texttt{You are an expert AI visual reasoning agent specialized in analyzing procedural and action videos.} \\
\\
\texttt{You will be provided with a video and a multiple-choice question. The video shows a user performing a task, but it is deliberately truncated (cut off) at a critical moment.} \\
\\
\texttt{Your task is to:} \\
\texttt{1. Thoroughly observe the video to understand the context, the user's progress, and the current situation.} \\
\texttt{2. Read the multiple-choice question, which asks about the specific situation, next step, or required action exactly at the moment the video ends.} \\
\texttt{3. Use your visual reasoning skills and general knowledge to deduce the correct answer based ONLY on this single video.} \\
\\
\texttt{OUTPUT CONSTRAINT:} \\
\texttt{You must output ONLY the exact letter corresponding to the correct option (e.g., A, B, C, D, or E).} \\
\texttt{Do NOT output any explanations, thinking processes, punctuation, or additional text. If the correct answer is option A, your entire response must be exactly: A}
\end{tcolorbox}

As described in the main text, this prompt was appended with the specific \texttt{Question} and \texttt{Options} for each evaluation instance. The use of strict output constraints (\textit{e.g., ``output ONLY the exact letter''}) ensures that Accuracy can be calculated via direct string matching, eliminating parsing ambiguities.

\section{Human Annotation Details}
\label{human_annotation}
To ensure high-fidelity labeling and the logical consistency of \OurBenchmark, we engaged a professional annotation team and implemented a multi-stage quality control pipeline. This section details the demographics, training mechanisms, and quality assurance protocols.

\subsection{Annotator Demographics and Qualifications}
A total of \textbf{99 annotators} participated in the project. We prioritized individuals with strong linguistic and technical backgrounds to handle the complex audio-visual instructions in the tutorials. The demographic breakdown is summarized in Table~\ref{tab:annotator_stats}.

\begin{table}[htbp]
  \centering
  \small
  \caption{Demographic and qualification statistics of the annotation team ($N=99$).}
  \vspace{5pt}
  \label{tab:annotator_stats}
  \begin{tabular}{llc}
    \toprule
    \textbf{Dimension} & \textbf{Category} & \textbf{Count / Percentage} \\
    \midrule
    \multirow{2}{*}{Age} & Under 25 years old & 47 (47.5\%) \\
    & Over 25 years old & 52 (52.5\%) \\
    \midrule
    \multirow{2}{*}{Education} & Bachelor's Degree & 74 (74.7\%) \\
    & Master's Degree & 25 (25.3\%) \\
    \midrule
    \multirow{2}{*}{Major} & CS / English Related & 58 (58.6\%) \\
    & Other Disciplines & 41 (41.4\%) \\
    \midrule
    \multirow{3}{*}{Language Proficiency} & CET 4 / 6 & 50 (50.5\%) \\
    & TEM-8 / IELTS $\ge$ 7.5 & 42 (42.4\%) \\
    & Overseas Study Experience & 7 (7.1\%) \\
    \midrule
    \multirow{2}{*}{Employment Status} & Full-time Professional & 93 (93.9\%) \\
    & Part-time / Intern & 6 (6.1\%) \\
    \bottomrule
  \end{tabular}
\end{table}

The majority of annotators (93.9\%) were full-time professionals, and nearly half (49.5\%) possessed elite-level English certifications or overseas experience, ensuring an accurate interpretation of nuanced verbal instructions in the tutorials.

\subsection{Training and Onboarding Mechanism}
To ensure all annotators fully mastered the Standard Operating Procedure (SOP), we implemented the following onboarding process:
\begin{itemize}[leftmargin=1.5em, nosep]
    \item \textbf{Systematic Training:} We organized a 3-hour comprehensive training session covering temporal calibration, question de-spoiling, and distractor engineering. Annotators engaged in real-time Q\&A sessions to align on all edge cases and logical rules.
    \item \textbf{Pilot Evaluation:} Every annotator underwent a pilot task evaluation. The initial pass rate was \textbf{90\%}. Annotators who failed were retrained and re-evaluated; we implemented a strict elimination policy for individuals with repeated failures to maintain data integrity.
\end{itemize}

\subsection{Annotation Platform and Workflow}
The project was conducted on a professional internal annotation platform. The workflow was designed as follows:
\begin{itemize}[leftmargin=1.5em, nosep]
    \item \textbf{Task Allocation:} All instances were uploaded to a centralized task pool. Annotators selected tasks randomly to prevent any systematic domain-specific bias.
    \item \textbf{Time Investment:} Each data instance required an average of \textbf{15--20 minutes} for complete refinement, reflecting the intensive nature of sub-second temporal calibration and distractor logic engineering.
\end{itemize}

\subsection{Quality Control and Conflict Resolution}
We implemented a multi-layered verification strategy to ensure the benchmark's reliability:
\begin{itemize}[leftmargin=1.5em, nosep]
    \item \textbf{Secondary Sampling:} A dedicated internal QA team performed a second-round audit on \textbf{20\%} of the completed samples.
    \item \textbf{Iterative Refinement:} Samples failing to meet the standard were returned for mandatory rework with specific improvement suggestions provided by the QA team.
    \item \textbf{Arbitration Mechanism:} In cases of logical ambiguity or disagreement, samples were flagged and resolved through an internal arbitration committee of senior researchers to ensure a unified and accurate ground truth.
\end{itemize}

\subsection{Ethics and Fair Compensation}
We strictly adhere to ethical labor practices. All annotators were compensated fairly, with an average wage exceeding the local market standard for professional data services. Annotators worked in regulated office environments with restricted hours to prevent fatigue and ensure consistent labeling quality.

\section{Statistical Reliability and Confidence Intervals}
\label{sec:ci}

In Table~\ref{tab:main_results} of the main paper, we report the \textbf{Wilson score intervals} at a 95\% confidence level for the accuracies achieved in the \textit{w/ Tutorial (Long)} setting. This section clarifies the methodology and interpretation of these statistical measures within the context of our benchmark.

\textbf{Methodology.} Given that our evaluation metric (Accuracy) is a binomial proportion, we employ the Wilson score interval to provide a robust estimate of the confidence range. Unlike the standard Wald interval (normal approximation), the Wilson interval is more reliable for finite sample sizes and remains accurate even when the observed proportions are near the extremes (0 or 1).

\textbf{Interpretation of Deterministic Results.} We emphasize that the evaluation on {\OurBenchmark{}} is designed to be \textbf{deterministic}. For all models, we use a greedy decoding strategy or a fixed sampling seed (e.g., Temperature = 0) to ensure that the model's response for a specific question is consistent across multiple runs. Therefore, the reported accuracy for the fixed test set does not exhibit experimental variance.

\textbf{Purpose of the Confidence Interval.} Although the results on our specific samples are deterministic, the confidence intervals serve as a measure of \textbf{statistical reliability} regarding the model's generalized performance. Since the {\OurBenchmark{}} test set consists of a finite number of samples ($N=2,261$), the confidence interval quantifies the uncertainty inherent in estimating a model’s ``true'' capability from a limited population. Specifically, the interval $[min, max]$ indicates that if we were to evaluate the model on an infinite sequence of similar episodic reasoning tasks, its true accuracy would fall within this range with 95\% probability. This provides readers with a clearer understanding of the significance of performance gaps between different architectures.

\newpage
\input{checklist.tex}

\end{document}

%% file: result_all.tex
\begin{table}[t]
  \centering
  \small
  \setlength{\tabcolsep}{3pt}
    \caption{Main results on the {\OurBenchmark{}} reported in Accuracy (\%). We compare model performance across three settings: (1) \textbf{w/o Tutorial} (baseline), (2) \textbf{w/ Tutorial} (immediate application), and (3) \textbf{w/ Tutorial (Long)} (episodic reasoning after a 2-hour distractor). $\Delta$ and $\Delta_L$ represent the performance gain attributed to tutorial knowledge over the baseline. Models are categorized into proprietary systems and open-source architectures, with the latter further divided into non-streaming and specialized streaming models. ``Frames'' denotes the total number of frames sampled per evaluation instance. Wilson 95\% confidence intervals are provided for w/ tutorial (Long) results.}
  \vspace{5pt}
  \label{tab:main_results}
  \begin{tabular*}{\textwidth}{@{\extracolsep{\fill}} l c c cc cc @{}}
    \toprule
    \multirow{2}{*}{\textbf{Model}} & \multirow{2}{*}{\textbf{Frames}} & \textbf{Baseline} & \multicolumn{2}{c}{\textbf{w/ Tutorial}} & \multicolumn{2}{c}{\textbf{w/ Tutorial (Long)}} \\
    \cmidrule(r){3-3} \cmidrule(lr){4-5} \cmidrule(l){6-7}
    & & \textbf{w/o Tut.} & \textbf{Acc.} & \textbf{$\Delta$} & \textbf{Acc.$_\text{CI}$} & \textbf{$\Delta_{L}$} \\
    \midrule

    \multicolumn{7}{c}{\textit{Proprietary Models}} \\ 
    \midrule
    Gemini 3.1 Pro \cite{gemini3_2025}  & 1fps & 45.45 & 74.92 & 29.47 & 59.45$_{[57.4, 61.5]}$ & 14.00 \\
    GPT-5.2 \cite{singh2025openaigpt5}         & 64 & 42.68 & 57.94 & 15.26 & 44.89$_{[42.9, 46.9]}$ & 2.21 \\
    GPT-4o \cite{hurst2024gpt4o}          & 64 & 35.16 & 52.83 & 17.67 & 37.64$_{[35.7, 39.7]}$ & 2.48 \\
    \midrule
    
    \multicolumn{7}{c}{\textit{Open-Source Models (Streaming)}} \\
    \midrule
    video-SALMONN-S (8B) \cite{vsalmonns}              & 1fps & 36.80 & 62.85 & 26.05 & 51.75$_{[49.7, 53.8]}$ & 14.95 \\
    StreamMem (8B) \cite{streammem} & 1fps & 37.02 & 64.62 & 27.60 & 43.26$_{[41.2, 45.3]}$ & 6.24 \\
    PEMF (8B) \cite{pemf}           & 1fps & 36.97 & 64.88 & 27.91 & 39.36$_{[37.4, 41.4]}$ & 2.39 \\

    \midrule
    \multicolumn{7}{c}{\textit{Open-Source Models (Non-Streaming)}} \\
    \midrule
    video-SALMONN 2+ (8B) \cite{vsalmonn2}            & 768 & 36.36 & 66.08 & 29.72 & 52.15$_{[50.1, 54.2]}$ & 15.79 \\
    video-SALMONN 2+ (8B) \cite{vsalmonn2}            & 64 & 37.64 & 63.64 & 26.00 & 47.24$_{[45.2, 49.3]}$ & 9.60 \\
    Qwen3-Omni (30B-A3B) \cite{Qwen3-Omni} & 64  & 38.46 & 54.14 & 15.68 & 40.79$_{[38.8, 42.8]}$ & 2.33 \\
    Qwen2.5-Omni (7B) \cite{Qwen2.5-Omni}           & 64  & 27.71 & 41.63 & 13.92 & 38.91$_{[36.9, 41.0]}$ & 11.2 \\
    VideoLLaMA3 (7B) \cite{damonlpsg2025videollama3}& 64  & 35.34 & 53.38 & 18.04 & 35.98$_{[34.0, 38.0]}$ & 0.64 \\
    Qwen3-VL (8B) \cite{qwen3vl}           & 64  & 33.44 & 49.21 & 15.77 & 34.72$_{[32.8, 36.7]}$ & 1.28 \\

    \bottomrule
  \end{tabular*}
\end{table}

%% file: task_result.tex
\begin{table}[tb]
\centering
\small
\caption{Detailed performance breakdown across fine-grained task categories on the {\OurBenchmark{}} benchmark, reported in Accuracy (\%). All values correspond to the \textbf{w/ Tutorial (Long)} setting, evaluating episodic reasoning after an extended temporal gap. Task abbreviations are as follows: \textbf{NSP} (Next Step Prediction), \textbf{SO} (Step Ordering), \textbf{SGP} (Subgoal Prediction), \textbf{PR} (Parameter Recall), \textbf{TC} (Tool Configuration), \textbf{SC} (Safety Check), \textbf{MD} (Mistake Detection), and \textbf{CB} (Conditional Branching).}
\vspace{5pt}
\label{tab:Task result}
\begin{adjustbox}{max width=\linewidth}
\begin{tabular}{lcccccccc} 
\toprule
\multicolumn{1}{c}{\multirow{1}{*}{\textbf{Models}}}  & \multicolumn{1}{c}{\textbf{NSP (\%)}} & \multicolumn{1}{c}{\textbf{SO (\%)}} & \multicolumn{1}{c}{\textbf{SGP (\%)}} & \multicolumn{1}{c}{\textbf{PR (\%)}} & \multicolumn{1}{c}{\textbf{TC (\%)}} & \multicolumn{1}{c}{\textbf{SC (\%)}} & \multicolumn{1}{c}{\textbf{MD (\%)}} & \multicolumn{1}{c}{\textbf{CB (\%)}} \\ 

\midrule
\multicolumn{9}{c}{\textit{Proprietary Models}} \\ 
\midrule

Gemini 3.1 Pro \cite{gemini3_2025} & 59.45 & 57.69 & 58.62 & 57.40 & 61.18 & 58.52 & 60.89 & 56.19 \\
GPT-5.2 \cite{singh2025openaigpt5} & 45.10 & 47.72 & 43.84 & 41.81 & 43.52 & 47.64 & 45.04 & 43.31 \\
GPT-4o \cite{hurst2024gpt4o} & 35.52 & 34.84 & 42.18 & 43.64 & 37.66 & 40.86 & 30.61 & 39.53  \\

\midrule
\multicolumn{9}{c}{\textit{Open-Source Models (Streaming)}} \\ 
\midrule
video-SALMONN-S (8B) \cite{vsalmonns} & 47.75 & 48.77 & 55.30 & 61.65 & 44.58 & 60.21 & 65.64 & 73.16 \\
StreamMem (8B) \cite{streammem} & 38.77 & 41.23 & 49.25 & 47.14 & 38.11 & 53.17 & 62.95 & 56.96 \\
PEMF (8B) \cite{pemf} & 36.71 & 37.43 & 48.24 & 39.86 & 34.86 & 46.20 & 46.14 & 59.50 \\

\midrule
\multicolumn{9}{c}{\textit{Open-Source Models (Non-Streaming)}} \\ 
\midrule
video-SALMONN2+ (8B) \cite{vsalmonn2} & 48.17 & 49.10 & 55.96 & 60.26 & 44.90 & 62.58 & 65.47 & 71.10 \\
Qwen3-Omni (30B-A3B) \cite{Qwen3-Omni} & 40.67 & 39.20 & 44.73 & 40.91 & 38.11 & 39.15 & 50.00 & 48.56 \\
Qwen2.5-Omni (7B) \cite{Qwen2.5-Omni} & 38.24 & 39.33 & 39.27 & 44.75 & 37.44 & 34.09 & 43.13 & 40.61 \\
VideoLLaMA3 (7B) \cite{damonlpsg2025videollama3} & 34.57 & 36.41 & 36.80 & 42.72 & 37.36 & 40.24 & 37.77 & 29.56 \\
Qwen3-VL (8B) \cite{qwen3vl} & 36.08 & 33.29 & 28.37 & 39.64 & 35.37 & 33.12 & 27.59 & 16.09 \\

\bottomrule

\end{tabular}
\end{adjustbox}
\end{table}

%% file: checklist.tex
\section*{NeurIPS Paper Checklist}

\begin{enumerate}

\item {\bf Claims}
    \item[] Question: Do the main claims made in the abstract and introduction accurately reflect the paper's contributions and scope?
    \item[] Answer:  \answerYes{} 
    \item[] Justification: We clearly listed the contributions in the introduction and in Sections 3 and 4, we demonstrate the contributions in detail.

\item {\bf Limitations}
    \item[] Question: Does the paper discuss the limitations of the work performed by the authors?
    \item[] Answer:  \answerYes{} 
    \item[] Justification: In section 5

\item {\bf Theory assumptions and proofs}
    \item[] Question: For each theoretical result, does the paper provide the full set of assumptions and a complete (and correct) proof?
    \item[] Answer:  \answerNA{}{} 
    \item[] Justification: N/A

    \item {\bf Experimental result reproducibility}
    \item[] Question: Does the paper fully disclose all the information needed to reproduce the main experimental results of the paper to the extent that it affects the main claims and/or conclusions of the paper (regardless of whether the code and data are provided or not)?
    \item[] Answer:  \answerYes{} 
    \item[] Justification: We provide metadata and full annotations in the link provided in the paper. We also provide result files from the models we evaluated.

\item {\bf Open access to data and code}
    \item[] Question: Does the paper provide open access to the data and code, with sufficient instructions to faithfully reproduce the main experimental results, as described in supplemental material?
    \item[] Answer:  \answerYes{} 
    \item[] Justification: The code for evaluation will be released in the provided link.

\item {\bf Experimental setting/details}
    \item[] Question: Does the paper specify all the training and test details (e.g., data splits, hyperparameters, how they were chosen, type of optimizer) necessary to understand the results?
    \item[] Answer:  \answerYes{} 
    \item[] Justification: In section 4.

\item {\bf Experiment statistical significance}
    \item[] Question: Does the paper report error bars suitably and correctly defined or other appropriate information about the statistical significance of the experiments?
    \item[] Answer:  \answerYes{} 
    \item[] Justification: In section 4

\item {\bf Experiments compute resources}
    \item[] Question: For each experiment, does the paper provide sufficient information on the computer resources (type of compute workers, memory, time of execution) needed to reproduce the experiments?
    \item[] Answer: \answerYes{} 
    \item[] Justification: In section 4.
    
\item {\bf Code of ethics}
    \item[] Question: Does the research conducted in the paper conform, in every respect, with the NeurIPS Code of Ethics \url{https://neurips.cc/public/EthicsGuidelines}?
    \item[] Answer: \answerYes{} 
    \item[] Justification: We have read the NeurIPS Code of Ethics and confirmed that our research conforms with it in every respect.

\item {\bf Broader impacts}
    \item[] Question: Does the paper discuss both potential positive societal impacts and negative societal impacts of the work performed?
    \item[] Answer: \answerNA{} 
    \item[] Justification: This paper is pure technical analysis.
    
\item {\bf Safeguards}
    \item[] Question: Does the paper describe safeguards that have been put in place for responsible release of data or models that have a high risk for misuse (e.g., pre-trained language models, image generators, or scraped datasets)?
    \item[] Answer: \answerYes{} 
    \item[] Justification: Sections 3 and 4 indicate full traceability of each sample and annotations.

\item {\bf Licenses for existing assets}
    \item[] Question: Are the creators or original owners of assets (e.g., code, data, models), used in the paper, properly credited and are the license and terms of use explicitly mentioned and properly respected?
    \item[] Answer: \answerYes{} 
    \item[] Justification: In section 4.

\item {\bf New assets}
    \item[] Question: Are new assets introduced in the paper well documented and is the documentation provided alongside the assets?
    \item[] Answer: \answerYes{} 
    \item[] Justification: In section 4 and in the repository link provided in the paper.

\item {\bf Crowdsourcing and research with human subjects}
    \item[] Question: For crowdsourcing experiments and research with human subjects, does the paper include the full text of instructions given to participants and screenshots, if applicable, as well as details about compensation (if any)? 
    \item[] Answer: \answerYes{} 
    \item[] Justification: We provide full instructions and screenshots in Appendix.

\item {\bf Institutional review board (IRB) approvals or equivalent for research with human subjects}
    \item[] Question: Does the paper describe potential risks incurred by study participants, whether such risks were disclosed to the subjects, and whether Institutional Review Board (IRB) approvals (or an equivalent approval/review based on the requirements of your country or institution) were obtained?
    \item[] Answer: \answerNA{} 
    \item[] Justification: This paper does not contain human subject study

\item {\bf Declaration of LLM usage}
    \item[] Question: Does the paper describe the usage of LLMs if it is an important, original, or non-standard component of the core methods in this research? Note that if the LLM is used only for writing, editing, or formatting purposes and does \emph{not} impact the core methodology, scientific rigor, or originality of the research, declaration is not required.
    \item[] Answer: \answerNA{} 
    \item[] Justification: LLM is only used for grammatical checks.

\end{enumerate}